\documentclass{article}

\usepackage{arxiv}

\usepackage[utf8]{inputenc} 
\usepackage[T1]{fontenc}    
\usepackage{hyperref}       
\usepackage{url}            
\usepackage{booktabs}       
\usepackage{amsfonts}       
\usepackage{nicefrac}       
\usepackage{microtype}      
\usepackage{lipsum}		
\usepackage{graphicx}
\usepackage{natbib}
\usepackage{doi}

\usepackage{graphicx}%
\usepackage{multirow}%
\usepackage{amsmath,amssymb,amsfonts}%
\usepackage{caption}
\usepackage{subcaption}
\usepackage{multirow}
\usepackage{booktabs}
\usepackage{graphicx}
\usepackage[table,xcdraw]{xcolor}
\usepackage[title]{appendix}%
\usepackage{xcolor}%
\usepackage{textcomp}%
\usepackage{manyfoot}%
\usepackage{booktabs}%
\usepackage{booktabs}
\usepackage{multirow}
\usepackage{algorithm}%
\usepackage{algorithmicx}%
\usepackage{algpseudocode}%
\usepackage{listings}%

\usepackage{relsize}
\usepackage{cleveref}
\usepackage[T1]{fontenc}

\usepackage{amssymb}
\usepackage{booktabs}
\usepackage{amsmath}

\title{Integrating Human Vision Perception in Vision Transformers for classifying waste items}

\author{{\hspace{1mm}Akshat Kishore Shrivastava} \\
	The Shri Ram School, Aravali\\
	Gurgaon, India, 122002 \\
	\texttt{akshat1509@icloud.com} \\
	\And
	{\hspace{1mm}Tapan Kumar Gandhi} \\
	Department of Electrical Engineering\\
	IIT-Delhi\\
	Delhi, India, 110016 \\
	\texttt{tgandhi@ee.iitd.ac.in} \\
}

\renewcommand{\shorttitle}{Human vision perception in ViT}

\hypersetup{
pdftitle={Integrating Human Vision Perception in Vision Transformers for classifying waste items},
pdfsubject={q-bio.NC, q-bio.QM},
pdfauthor={Akshat Kishore Shrivastava, Tapan Kumar Gandhi},
pdfkeywords={waste classification, artificial intelligence, vision transformers, human vision},
}

\begin{document}
\maketitle

\begin{abstract}
	In this paper, we propose an novel methodology aimed at simulating the learning phenomenon of nystagmus through the application of differential blurring on datasets. Nystagmus is a biological phenomenon that influences human vision throughout life, notably by diminishing head shake from infancy to adulthood. Leveraging this concept, we address the  issue of waste classification, a pressing global concern. The proposed framework comprises two modules, with the second module closely resembling the original Vision Transformer, a state-of-the-art model model in classification tasks. The primary motivation behind our approach is to enhance the model's precision and adaptability, mirroring the real-world conditions that the human visual system undergoes. This novel methodology surpasses the standard Vision Transformer model in waste classification tasks, exhibiting an improvement with a margin of 2\%. This improvement underscores the potential of our methodology in improving model precision by drawing inspiration from human vision perception. Further research in the proposed methodology could yield greater performance results, and can be extrapolated to other global issues. 
\end{abstract}


\section{Introduction}
\label{sec:sample1}
In the stage of infancy, children have multiple forms of visual simulations and do not possess clear vision. This is due to the phenomenon of nystagmus. One concept of human vision perception which is not accounted for by conventional architectures is that of such nystagmus. Nystagmus \cite{nihNystagmusTypes}, by definition, is the involuntary shaking of the head which leads to decrease in visibility, or blurring of visual senses. Humans are known to gain clearer vision throughout the years of growth- they are known to have a blurred vision at birth which gets clearer with time, allowing us to eventually perceive the world. The amount of head shake induced visual blurring is reduced as a person ages. In this work, we aim to aim to simulate this phenomenon of nystagmus in the training of models, so that it better mimics the methods of human vision perception in models. \par
Even though such a phenomenon is known to be a part of the human visual perceptiveness, there is no tool for quantifying this phenomenon, and no efforts to correlate the idea of such visual blurriness to that of artificial intelligence models. Since such models are inspired by the human visual system, the idea behind the proposed paper is to overcome these claims while presenting a novel method for quantifying nystagmus and using the data from nystagmus to simulate human visual learning via differential blurring. \par
We apply this concept to perform the task of waste classification. Only 13.5\% percent of the world's garbage, according to recent estimates, is recycled, and 33\% percent is dumped in the open without being classified \cite{Kaza2018}. Soil contamination, surface and ground water pollution, greenhouse gas emissions, and decreased crop productivity are common risks related to openly dumping unsorted garbage \cite{book}. However, only 17.4\% percent of electronic garbage generated worldwide is collected and recycled, at a cost of over 57 billion US dollars (USD). The Ellen MacArthur Foundation, which asserted that 32 types of plastic packaging are not collected, estimated the economic cost to be between 80 billion USD and 120 billion USD. By 2050, the rate of expansion of the world's garbage is predicted to surpass that of the population \cite{LI2016333}, which will have major consequences for the ecological balance as well as for human well-being and global sustainable development. This necessitates the creation of tools that enhance waste management automation and mitigate the lack of awareness. \par
The rapid development of computer vision and artificial intelligence models has made automatic recognition and detection of waste from photos a popular alternative to human waste sorting \cite{article}. To increase the precision of autonomous waste classification, numerous machine learning techniques have been developed \cite{article1, article2, DBLP:journals/corr/abs-1710-10741}. However, deep neural networks \cite{article10}, particularly convolutional neural networks (CNN), have recently demonstrated their superior ability to learn from pre-existing data, producing accurate results in image categorization \cite{article1, article3, DBLP:journals/corr/abs-1802-05751, DBLP:journals/corr/abs-1904-10509}. However, an architecture known as the Vision Transformer \cite{DBLP:journals/corr/abs-2010-11929} proposed a methodology which can completely eliminate the need for convolutions, therefore reducing computation times.  \par
Recently, in the field of computer vision, transformer-based architectures have proved to be an accurate alternative to neural networks. \cite{DBLP:journals/corr/VaswaniSPUJGKP17}, which is becoming a more and more common option for both study and practise. Initially, transformers were commonly used in the field of natural language processing (NLP) . \cite{DBLP:journals/corr/abs-2005-12872, DBLP:journals/corr/abs-1810-04805}. Recently, the vision transformer architecture was proposed to transfer the transformer architecture to vision related tasks \cite{DBLP:journals/corr/abs-2005-12872, DBLP:journals/corr/abs-2010-11929}. Currently, Vision Transformers are considered to be one of the state-of-the-art classifiers in image classification.  Vision Transformers have gained massive popularity with the self-attention mechanism totally removing the need for any convolutional networks \cite{Wang_2018_CVPR}. 

Overall, in this paper, we make the following contributions:
\begin{itemize}
  \item Simulate the biological learning effect of nystagmus in the training of artificial intelligence models. 
  \item Apply this to a task of waste classification, an impending global issue, through the usage of a vision transformer architecture.
\end{itemize}

\section{Related Work}
\label{sec:sample1}
\textbf{Vision Transformers} were introduced by Dosovitskiy et al. \cite{DBLP:journals/corr/abs-2010-11929}, which build upon the initial transformer model built for NLP by Vaswani et al. \cite{DBLP:journals/corr/VaswaniSPUJGKP17, DBLP:journals/corr/abs-2005-14165, britz-etal-2017-massive}. The idea of eliminating or reducing convolutions for image classification tasks has been explored in various papers, through the usage of local self-attention mechanism. Parmar et al. \cite{DBLP:journals/corr/abs-1802-05751}, Hu et al. \cite{DBLP:journals/corr/abs-1904-11491}; Ramachandran et al. \cite{DBLP:journals/corr/abs-1906-05909}; Zhao et al. \cite{DBLP:journals/corr/abs-2004-13621} applied the self-attention locally, proposing architectures to eliminate convolutions altogether or partially. Cordonnier et al. \cite{DBLP:journals/corr/abs-1911-03584} used a methodology to vectorise, or “patchify” images, allowing them to be passed into a self-attention architecture. \par
\textbf{Waste classification through the usage of artificial intelligence} has been a major topic of research due to the increasing waste crisis across the globe. There are multiple research works which use both state-of-the-art and custom convolutional networks to complete waste classification tasks. Abdulfattah E. al. \cite{9493430} use multiple state-of-the-art CNN’s, implementing DenseNet121 to achieve the task with highest accuracy. Wang et al. \cite{article4}, who tested a fine-tuned VGGNet-19 on a self-composed trash data set of 69,737 photos, achieved 86.19\% accuracy. On two datasets, containing 372 and 72 images, respectively, Dewulf \cite{phdthesis} examined the performance of four state-of-the-art CNN architectures: AlexNet \cite{article5}, VGGNet \cite{article6}, GooLeNet \cite{DBLP:journals/corr/SzegedyLJSRAEVR14}, and InceptionNet \cite{DBLP:journals/corr/SzegedyVISW15}. The best accuracy was produced by VGGNet and Inception-v3, with 91.40\% and 93.06\%, respectively. Mallikarjuna et al. \cite{inproceedings} obtained 90\%  accuracy with a four-layer CNN. A hybrid CNN-ELM model with best accuracy between 91\% and 93\% was implemented by Junjie et al. \cite{teh2020household} to classify waste items. To classify similar waste data, Xie et al. \cite{DBLP:journals/corr/XieGDTH16} built a CNN model using the combined Residual Transformations Network (ResNeXt). The VN-trash dataset, which consists of 5904 photos divided into three distinct waste classes (organic, inorganic, and medical), as well as the TrashNet dataset, which comprises 2527 images divided into six waste classes (glass, paper, cardboard, plastic, metal, and garbage), were used to evaluate the model. On these datasets, their model produced accuracy of 98\% and 94\%, respectively. With VGGNet-16, Buelaevanzalina \cite{kumar2021efficient} achieved 83\% accuracy, whilst Kusrini \cite{andhy2021waste} obtained an f-score ranging from 69 to 82\% with YOLOv4, using 3870 photos from the classes of glass, metal, paper, and plastic being included in a multi-class dataset. With waste data of 2527 photos, Castellano et al.'s \cite{inproceedings2} evaluation of VGGNet-16 obtained 85\% accuracy. Alonso et al. \cite{article7} used 4 classes of paper, plastic, organic trash, and glass to assess an unnamed CNN architecture with 3600 self-obtained waste photos. They obtained an f-score of between 59\% and 75\% for each class. Srinilta and Kanharattanachai \cite{Srinilta2019MunicipalSW} used 4 categories of waste: compostable, hazardous, general and recyclable. In order to facilitate waste classification, this data was utilised to train four state-of-the-art CNN architectures, including VGGNet \cite{article6}, ResNet-50 \cite{DBLP:journals/corr/HeZRS15}, MobileNet-v2 \cite{8578572}, and DenseNet-121 \cite{DBLP:journals/corr/HuangLW16a}. \par
In various articles, the \textbf{idea of nystagmus} and relation to computer vision have been examined. Mehrdad Sangi et al. \cite{article8} proposed a methodology entailing tracking the head, stabilising the head, cropping the eye region, detecting the centre of the pupil, and tracking the limbus edge, which produced an eye displacement and velocity signal. Nicolas Huynh Thien et al. \cite{Thien2012HorizontalGN} proposed an algorithm for estimating alcohol intoxication by detecting nystagmus in subjects. In the study by Li et al. \cite{s23031592}, a method for detecting nystagmus based on deep learning and using image processing techniques has been proposed as a tool for identifying nystagmus. Tomasz Pander et al. \cite{article9} have developed a novel method for detecting saccadic eye movements and Jason Turuwhenua et al. \cite{article11} used the optical flow of the limbus to identify nystagmus. \par
The idea of \textbf{Human Vision integration} in artificial intelligence models is greatly inspired by Haritosh et al \cite{9029047} who used a set amount of blurring in general classification tasks. \par

\section{Methods}
\label{sec:sample1}
\subsection{Module 1: Nystagmus Simulation}
Let the input matrix be $X \in \mathbb{R} ^ {m \times n}$ such that: 
\[ X = \{x_i : x_i \in \mathbb{R} ^ {n \times 1}, i \in {1 ... m}\} \] \par
Where $m$ is the size of the input sample space and $n$ is the size of the input feature space, or the number of images. 
Now, the phenomenon of nystagmus is simulated through the use of $k$ levels of Gaussian Blur, such that the model first sees the most blurred image and sees the least blurred image last, thereby simulating the gradual human vision perception. \par
To achieve this, the dataset is divided into $k$ subsets such at:
\begin{align*}
    \bigcup_{i=1}^{k}\Tilde{\mathbf{X}}_i = \mathbf{X} \quad \quad \text{and} \quad \quad \bigcap_{i=1}^{k}\Tilde{\mathbf{X}}_i = \phi
\end{align*}

To obtain $k$ levels of Gaussian blur, we change both $y$, the kernel size, as well as $\sigma$, the variance of the Gaussian distribution used to blur the image. These parameters are used to form a Gaussian Kernel, $G(x,y)$, as given below:
\[
G(i, j) = \frac{1}{2\pi\sigma^2} e^{-\frac{i^2 + j^2}{2\sigma^2}}
\]
\par
This kernel is used to blur the initial image $I(x,y)$ to get a new image matrix, $B(x,y)$.
\[
B(x, y) = \sum_{i=-y}^{y} \sum_{j=-y}^{y} I(x+i, y+j) \cdot G(i, j)
\]
\par
These are the equations used to apply a fixed amount of blur to a singular image. In our framework, we need to apply $k$ levels of blur to $n$ image vectors. To do this, we form a series of $y$ and $\sigma$ using an integer value $b$ such that $b \in \{ 0 ... k-1 \}$. The $n$ images are split into $n // k$ groups. The blur levels, $b$ are used to generate different values for $y$ and $\sigma$ through the equations below:
\[
y =  2b + 1
\]
\[
\sigma = b \cdot 0.3 + 0.5
\]
\par
In this work, we use a linear variation to change the values of $y$ and $\sigma$, gradually blurring images in the full dataset. 

\begin{figure}
    \centering
    \includegraphics[width=\textwidth]{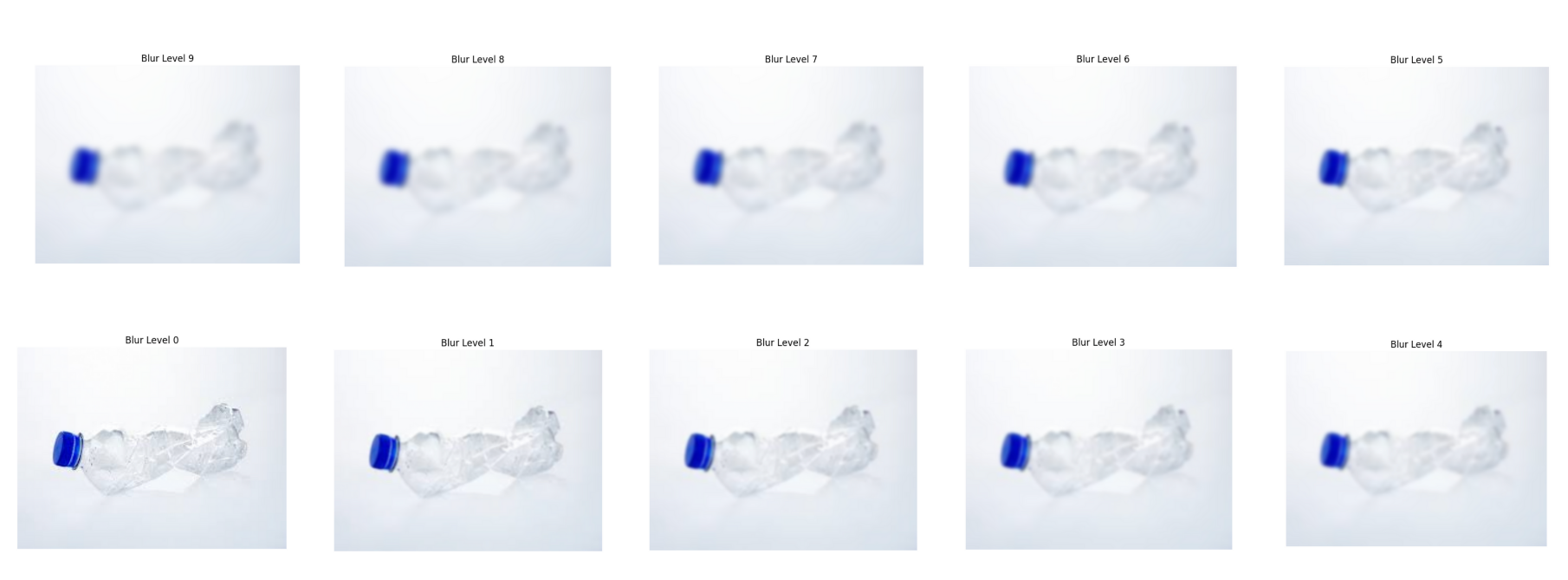}
    \caption{A visual representation of nystagmus simulation on a single image.}
\end{figure}
 
\subsection{Module 2: Vision Transformer}
\textbf{Image tokenisation}: Formally, each image is initially a 3D vector in a 2D representation. That is, each image is initially $x \in \mathbb{R}^{H\times W\times C}$, where $(H,W)$ are the dimensions of the image, $C$ represents the colour channels. 

This image is reshaped to $x_p \in \mathbb{R}^{N\times (P^2C)}$, where $(P,P)$ is the resolution of each patch, and $N = HC/P^2$. This new vector represents each image: each row of this matrix is a particular patch of size 16x16, 14x14 or any other specified size. 

Each 1D vector, which represents a single patch, is converted to a linear vector of size $D$, using a linear projection, where $D$ is the constant latent vector size for the transformer.

In addition to these projected patches, a \texttt{[class]} token, a trainable embedding, is also appended to these patches. It is the state of this token at output which gives the relevant image classification. The \texttt{[class]} token is the appended as the first row of each image sequence. At the output state, there will be an output vector for each feature row. However, the classification is performed using the state of the \texttt{[class]} token since it consists of information from every other patch, due to the self-attention mechanism described in a later section. 

Additionally, a positional embedding is added to the image representation. These embeddings express where the patch was initially present in the image. This can be done by simply assigning values such as 1, 2, 3... to each patch in an image. Another method for embedding positional information is the use of $sin$ and $cos$ function, making the embeddings smooth and unique, to integrate relative positional information based on frequency, as expressed in the formulae below, depending on if $j$ is even or odd respectively: \par
\[ p_{i,j} = sin \left( \frac{i}{10,000^\frac{j}{d}} \right)\]
\[ p_{i,j} = cos \left( \frac{i}{10,000^\frac{j-1}{d}} \right)\]
\par
As is evident from these expressions, the wavelengths of the sinusoids range from $2\pi$ to $10,000.2\pi$. Each row vector is a sinusoidal series whose frequency increases according to geometric progression i.e. within each row the frequency monotonically increases. Subsequently, each row represents the position encoding of a single discrete position. 

Now, the matrix obtained in through the previous equations is passed to the vision transformer \cite{DBLP:journals/corr/abs-2010-11929}. The input matrix, therefore, consists of  two-dimensional arrays. Formally, the transformer takes the following into the first layer:
\[ z_0 = [x_{class}; x_p^1E; x_p^2E; ... ; x_p^NE] + p \] 
\par
This equations represents both the patch and position embeddings, as well as the class token. Each patch vector is a vector of size $D$. The vector $E$ is used to linearly map the patch vectors to this required patch size $D$. 

\textbf{Multi-head self attention}: The first block in the architecture performs the function of multi-head self attention: 
\[ z'_l = MSA(LN(z_{l-1})) + z_{l-1} \]

The inputs are first normalised \cite{article13} and then passed to the self-attention layer. The addition with $z_{l-1}$ represents the residual connection with previous blocks. 

This module enables the model to evaluate the significance of various patches in relation to other patches of the same image. Based on this relation, each self-attention head calculates attention weights for each patch. The amount that each patch should contribute to the final representation of the image is determined by these attention weights. 

This layer calculates a set of attention ratings between each pair of patch embeddings, utilising trainable parameters to assess the relative significance of each patch. The weighted total of the patch embeddings is calculated using the attention scores, where the weights represent the significance of each patch. The MSA may capture both local and global dependencies in an image by employing multiple attention heads and then appending them. 

Mathematically, let the MSA blocks take in the input vector $x$. This $x$ is linearly projected into 3 matrices using trainable weights, called $Q$, $K$ and $V$ or the Query, Key and Value matrices. Using this matrices, the attention scores are received as below:
\[ A = softmax \left( \frac{QK^T}{\sqrt{d_k}} \right)  \]
\par
Here, $QK^T$ is a dot product, which is scaled by division with $\sqrt d_K$ to prevent very large values, and then passed to the softmax function to ensure that higher values near 1 are assigned to more relevant positions of the sequence. The dot product of $Q$ and $K^T$ creates a matrix which represents the similarity of each patch with every other patch. In effect, a particular row of the dot product matrix represents the similarity score of that patch with every other patch. These weights are used to obtain a re-weighted patch representation by re-weighting $V$: \par
\[Z =AV\]
\par
$Z$ is a weighted sum of $V$, where the values in the matrix $A$ determine the importance of each value in $A$ to the final weighted value in $Z$. \par
In a multi-head self attention module, $k$ self-attention heads run in parallel. Since there are multiple attention heads, all outputs are concatenated and projected at the end of computation. Initially, the input sequence is divided into $k$ sub-arrays to compute self-attention and concatenated to give an output equal to the initial input size. These weighted values are the output for the MSA block.
\par

\textbf{MLP and Outputs}: 
The output of the self-attention layers are passed through MLP blocks.
\[ z_l = MLP(LN(z'_l)) +z'_l \]
\par
Two types of layers, a linear layer and a Gaussian Error Linear Units (GELU), are present in the MLP module. These layers first convert the input to a higher dimensionality, apply a Gaussian Error Linear Unit (GELU), and then convert the sequences back to the $D$ sized patch embeddings in order to keep output sizes consistent. The final patch embedding that have undergone the MLP layer's transformation is created by adding the output of the MLP layer to the input patch embedding after MSA. Such MLP modules help the model capture complex patterns and relationships between the patch embeddings by introducing a non-linear function. 

The output of the transformer is received through the normalisation of the final state of the \texttt{[class]} token:
\[ y = LN(z_L^0) \]
\par

\begin{figure}
    \centering
    \includegraphics[width=\textwidth]{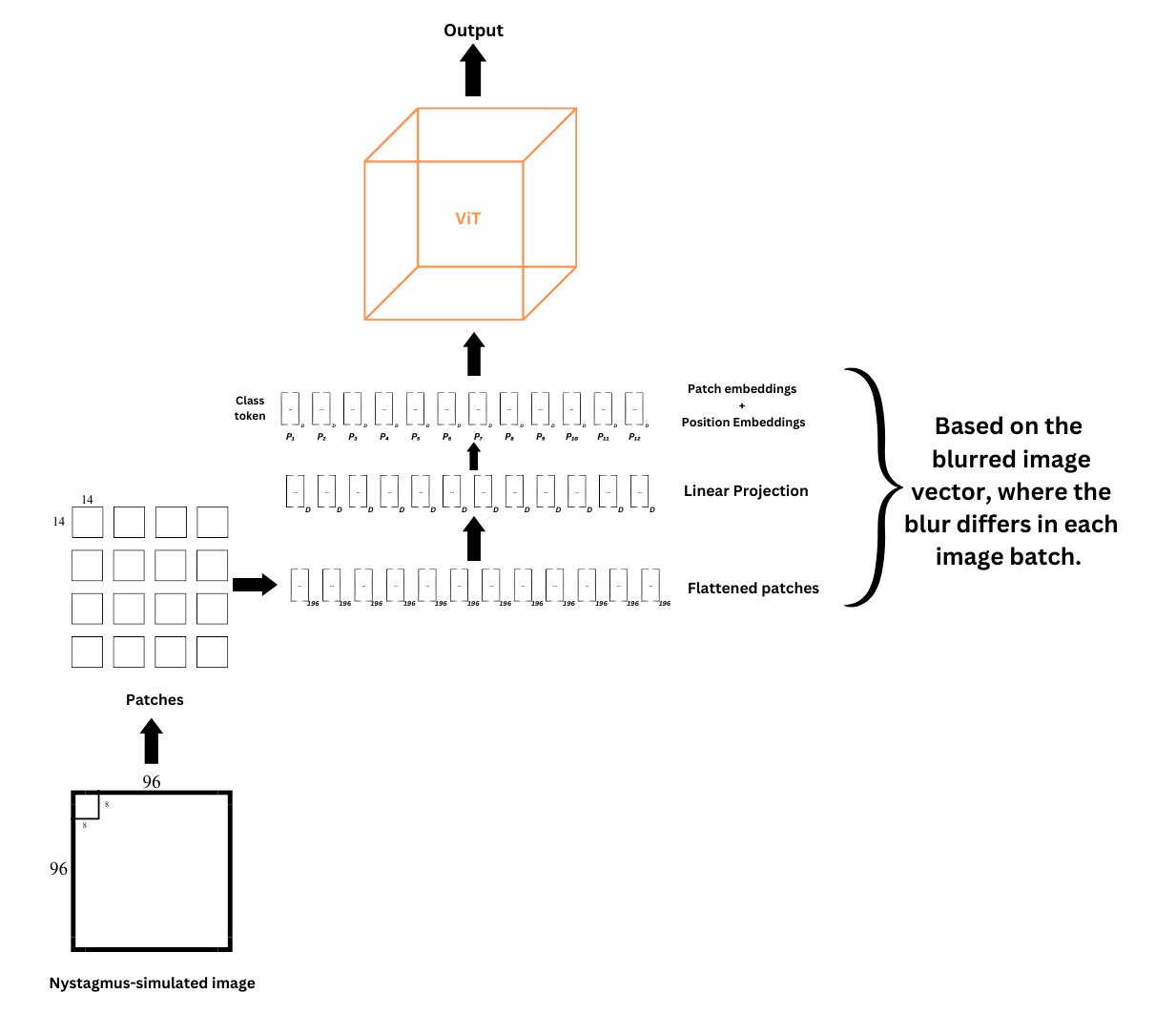}
    \caption{A graphical representation of the HP-ViT framework.}
\end{figure}

\section{Results}
\subsection{Experimental Setup}
The train-matrix is $X \in \mathbb{R} ^ {m \times n}$, where m and n is the sample space of the train-matrix, comprises of 22,564 samples, and image feature space with $224 \times 224 \times 3$ features. There are 2513 test images of the same image dimensionality. In addition to this binary data set, we also test the proposed methodology on trashnet, which has 2527 images from 6 classes, with the following distribution: 501 glass, 594 paper, 403 cardboard, 482 plastic, 410 metal and 137 trash. This data is split so as to obtain 252 test images. 

We use a vision transformer with $D = 32$, $k = 16$, and $nBlocks = 4$. The architecture is built keeping in mind both the model performance as well as optimisation. In terms of simulating nystagmus, we use 10 levels of blur, generating a set of 10 values for both $y$ and $\sigma$. The train data is similarly split into 10 groups, each a different amount of blur corresponding to the respective $y$ and $\sigma$ values.  

The `Human Perception - ViT' framework is implemented using the PyTorch library of the Python programming language. Numpy and Pandas libraries are used for mathematical computation and data-related tasks, respectively.

\begin{table}[!htpb]
\centering
\scriptsize
\caption{Dataset breakup for the binary dataset}
\label{metrics}
\begin{tabular}{@{}lrr@{}}
\toprule
 & \textbf{Non-biodegradable wastes} & \textbf{Biodegradable wastes} \\
 \midrule
Train Matrix & 9999 & 12,600  \\
Test Matrix & 1112 & 1401  \\
\bottomrule
\end{tabular}
\end{table}

\begin{table}[!htpb]
\centering
\scriptsize
\caption{Dataset breakup for trashnet}
\label{metrics}
\begin{tabular}{@{}lrrrrrr@{}}
\toprule
 & \textbf{Cardboard} & \textbf{Glass} & \textbf{Metal} & \textbf{Paper} & \textbf{Plastic} & \textbf{Trash} \\
 \midrule
Train Matrix & 363 & 451 & 369 & 535 & 434 & 123 \\
Test Matrix & 40 & 50 & 41 & 59 & 48 & 14 \\
\bottomrule
\end{tabular}
\end{table}

\subsection{Classification Performance}
The novel Human Perception Based ViT architecture is compared to the standard ViT architecture which it seeks to improve. It is to be noted that while the models are trained on the train-data, the classification performance of the models is tested on the test-matrix, comprising 2513 unseen test images. Various performance metrics such as accuracy, precision, recall, and F1-score are computed and compared in the table below. In addition, the figure shows the classification confusion matrices for the models. The HP-ViT model reaches an accuracy of 90.61\% while the ViT model attains an accuracy of 88.70\%.  \par

\begin{figure}[!htpb]
     \centering
     \begin{subfigure}[b]{0.65\textwidth}
         \centering
         \includegraphics[width=\textwidth]{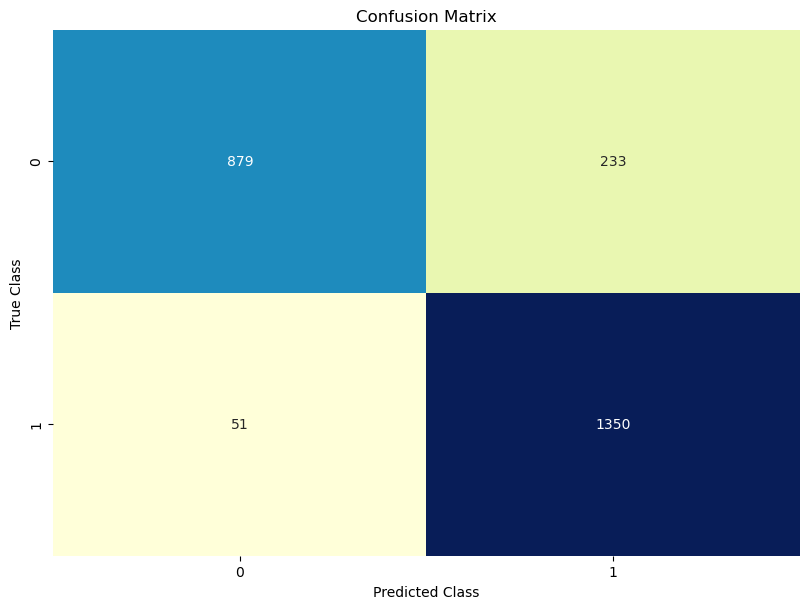}
         \caption{}
         \label{cm1}
     \end{subfigure}
     \\
     \begin{subfigure}[b]{0.65\textwidth}
         \centering
         \includegraphics[width=\textwidth]{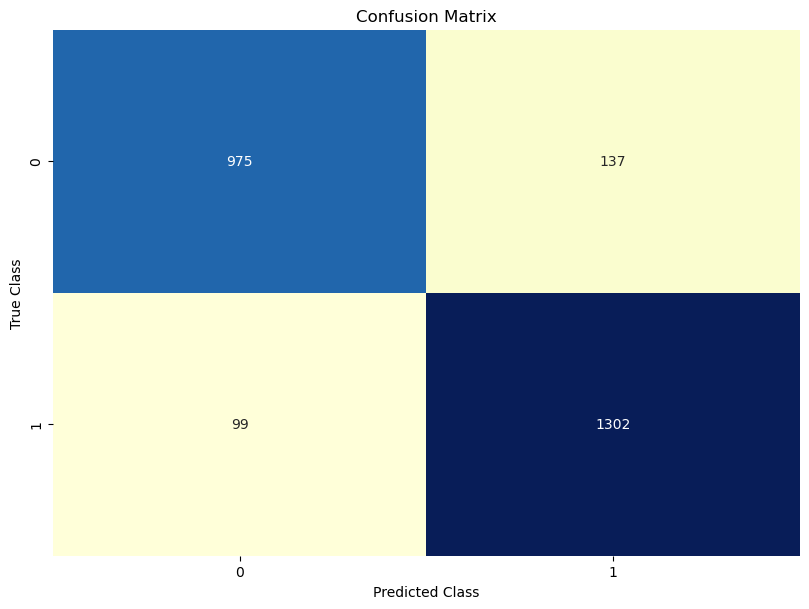}
         \caption{}
         \label{cm2}
     \end{subfigure}
         \caption{Confusion matrices for given models; For the ViT architecture (\Cref{cm1}); and HP-ViT architecture (\Cref{cm2}) on the binary dataset.}
\end{figure}

In addition to this comparison for a binary dataset, we run the model on the trashnet dataset, consisting of more classes and significantly lesser images. In this dataset, there are 252 test images. The HP-ViT architecture reaches an AUROC score of 0.87, while the ViT architecture reaches an AUROC score of 0.84 on this dataset. 

\begin{figure}[!htpb]
     \centering
     \begin{subfigure}[b]{0.49\textwidth}
         \centering
         \includegraphics[width=.99\linewidth]{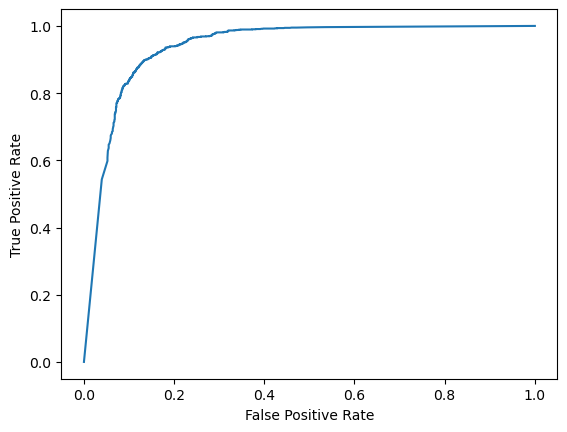}
         \caption{}
         \label{r1}
     \end{subfigure}
     \begin{subfigure}[b]{0.49\textwidth}
         \centering
         \includegraphics[width=.99\linewidth]{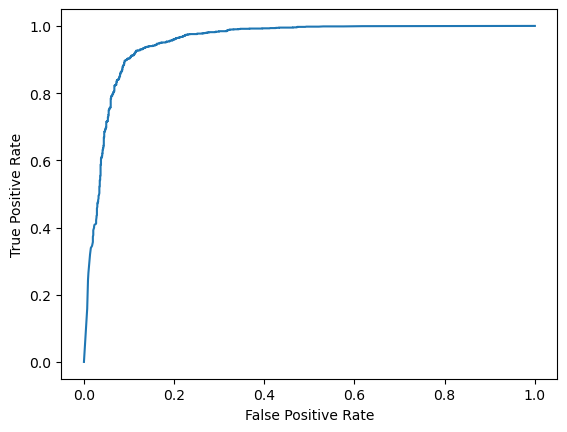}
         \caption{}
         \label{r2}
     \end{subfigure}
     \\
     \begin{subfigure}[b]{0.49\textwidth}
         \centering
         \includegraphics[width=.99\linewidth]{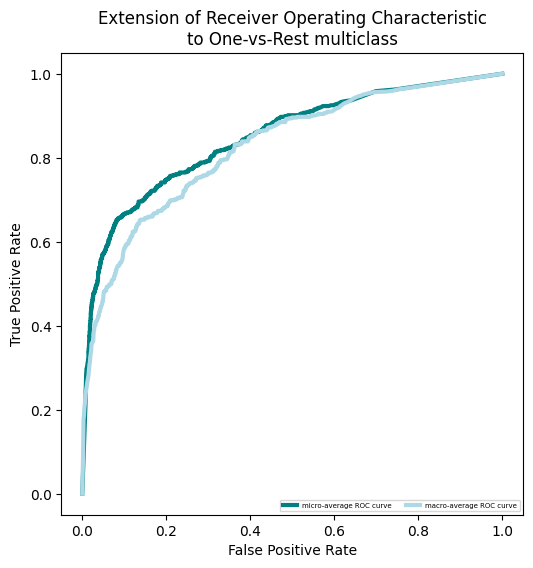}
         \caption{}
         \label{r3}
     \end{subfigure}
     \begin{subfigure}[b]{0.49\textwidth}
         \centering
         \includegraphics[width=.99\linewidth]{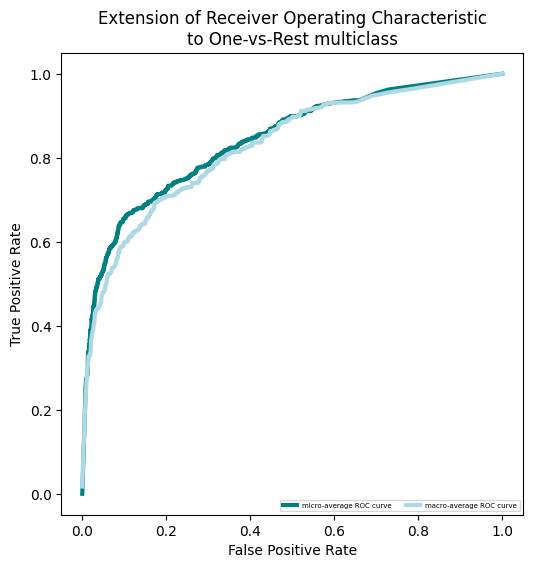}
         \caption{}
         \label{r4}
     \end{subfigure}
         \caption{The receiver operating characteristic curves for given models: for the ViT architecture on the binary dataset (\Cref{r1}) with AUROC=0.94; the HP-ViT architecture on the binary dataset (\Cref{r2}) with AUROC=0.96; the ViT architecture on the trashnet dataset (\Cref{r3}) with AUROC=0.84; and the HP-ViT architecture on the trashnet dataset (\Cref{r4}) with AUROC=0.87}
\end{figure}

The following performance measures are used to measure the classification performance of the model on the binary dataset:
\[Accuracy = \frac{TP+TN}{TP+TN+FP+FN} \]
\[Precision = \frac{TP}{TP+FP} \]
\[ Recall = \frac{TP}{TP+FN} \]
\[ F1 = \frac{2*Precision*Recall}{Precision+Recall} = \frac{2*TP}{2*TP+FP+FN} \]
\par

\begin{table}[!htpb]
\centering
\scriptsize
\caption{Comparison of various performance metrics of the HP-ViT methodology with the ViT model on the binary dataset}
\label{metrics}
\begin{tabular}{@{}lrrrr@{}}
\toprule
\textbf{Model} & \textbf{Accuracy} & \textbf{Precision} & \textbf{Recall} & \textbf{F1-Score} \\ \midrule
HP-ViT & 0.9061 & 0.91 & 0.90 & 0.90 \\
ViT & 0.8870 & 0.90 & 0.88 & 0.88 \\
\bottomrule
\end{tabular}
\end{table}

\section{Conclusion}
In this work, we propose a novel methodology to process data sets before passing them into any form of artificial intelligence models. In this case, we apply this methodology to the task of waste classification, on a total of 2 different data sets, passing the modified data sets into a vision transformer architecture. \par
The proposed methodology surpasses the standard ViT, and also gains more accuracy than other state-of-the-art works on these data sets. In the future, this principle of human vision integration in data sets can be tested on a variety of other data sets. Moreover, the mathematical approach to varying the Gaussian blur parameters can be changed to simulate another forms of growth, such as exponential.

\bibliographystyle{unsrtnat}
\bibliography{cas-refs}  

\begin{thebibliography}{48}
\providecommand{\natexlab}[1]{#1}
\providecommand{\url}[1]{\texttt{#1}}
\expandafter\ifx\csname urlstyle\endcsname\relax
  \providecommand{\doi}[1]{doi: #1}\else
  \providecommand{\doi}{doi: \begingroup \urlstyle{rm}\Url}\fi

\bibitem[Sekhon~RK(2023)]{nihNystagmusTypes}
Deibel~JP Sekhon~RK, Rocha Cabrero~F.
\newblock Nystagmus types.
\newblock StatPearls [Internet]. Treasure Island (FL): StatPearls Publishing, 2023.
\newblock [Accessed 04-09-2023].

\bibitem[Kaza et~al.(2018)Kaza, Yao, Bhada-Tata, and Woerden]{Kaza2018}
Silpa Kaza, Lisa~C. Yao, Perinaz Bhada-Tata, and Frank~Van Woerden.
\newblock \emph{What a Waste 2.0: A Global Snapshot of Solid Waste Management to 2050}.
\newblock Washington, {DC}: World Bank, August 2018.
\newblock \doi{10.1596/978-1-4648-1329-0}.
\newblock URL \url{https://doi.org/10.1596/978-1-4648-1329-0}.

\bibitem[Forti et~al.(2020)Forti, Baldé, Kuehr, and Bel]{book}
Vanessa Forti, Cornelis Baldé, Ruediger Kuehr, and Garam Bel.
\newblock \emph{The Global E-waste Monitor 2020. Quantities, flows, and the circular economy potential}.
\newblock 07 2020.
\newblock ISBN 978-92-808-9114-0.

\bibitem[LI et~al.(2016)LI, TSE, and FOK]{LI2016333}
W.C. LI, H.F. TSE, and L.~FOK.
\newblock Plastic waste in the marine environment: A review of sources, occurrence and effects.
\newblock \emph{Science of The Total Environment}, 566-567:\penalty0 333--349, 2016.
\newblock ISSN 0048-9697.
\newblock \doi{https://doi.org/10.1016/j.scitotenv.2016.05.084}.

\bibitem[Xia et~al.(2021)Xia, Jiang, Chen, and Zhao]{article}
Wanjun Xia, Yanping Jiang, Xiaohong Chen, and Rui Zhao.
\newblock Application of machine learning algorithms in municipal solid waste management: A mini review.
\newblock \emph{Waste Management `I\&' Research: The Journal for a Sustainable Circular Economy}, 40:\penalty0 0734242X2110337, 07 2021.
\newblock \doi{10.1177/0734242X211033716}.

\bibitem[Ye et~al.(2019)Ye, Yang, Zhong, Tu, Jia, and Wang]{article1}
Zhiping Ye, Jiaqian Yang, Na~Zhong, Xin Tu, Jining Jia, and Jiade Wang.
\newblock Tackle environmental challenges in pollution controls using artificial intelligence: A review.
\newblock \emph{Science of The Total Environment}, 699:\penalty0 134279, 09 2019.
\newblock \doi{10.1016/j.scitotenv.2019.134279}.

\bibitem[LeCun et~al.(2015)LeCun, Bengio, and Hinton]{article2}
Yann LeCun, Y.~Bengio, and Geoffrey Hinton.
\newblock Deep learning.
\newblock \emph{Nature}, 521:\penalty0 436--44, 05 2015.
\newblock \doi{10.1038/nature14539}.

\bibitem[Sun et~al.(2017)Sun, Xue, and Zhang]{DBLP:journals/corr/abs-1710-10741}
Yanan Sun, Bing Xue, and Mengjie Zhang.
\newblock Evolving deep convolutional neural networks for image classification.
\newblock \emph{CoRR}, abs/1710.10741, 2017.
\newblock URL \url{http://arxiv.org/abs/1710.10741}.

\bibitem[Rawat and Wang(2017)]{article10}
Waseem Rawat and Zenghui Wang.
\newblock Deep convolutional neural networks for image classification: A comprehensive review.
\newblock \emph{Neural Computation}, 29:\penalty0 1--98, 06 2017.
\newblock \doi{10.1162/NECO_a_00990}.

\bibitem[Liang and Gu(2021)]{article3}
Shuang Liang and Yu~Gu.
\newblock A deep convolutional neural network to simultaneously localize and recognize waste types in images.
\newblock \emph{Waste Management}, 126:\penalty0 247--257, 05 2021.
\newblock \doi{10.1016/j.wasman.2021.03.017}.

\bibitem[Parmar et~al.(2018)Parmar, Vaswani, Uszkoreit, Kaiser, Shazeer, and Ku]{DBLP:journals/corr/abs-1802-05751}
Niki Parmar, Ashish Vaswani, Jakob Uszkoreit, Lukasz Kaiser, Noam Shazeer, and Alexander Ku.
\newblock Image transformer.
\newblock \emph{CoRR}, abs/1802.05751, 2018.
\newblock URL \url{http://arxiv.org/abs/1802.05751}.

\bibitem[Child et~al.(2019)Child, Gray, Radford, and Sutskever]{DBLP:journals/corr/abs-1904-10509}
Rewon Child, Scott Gray, Alec Radford, and Ilya Sutskever.
\newblock Generating long sequences with sparse transformers.
\newblock \emph{CoRR}, abs/1904.10509, 2019.
\newblock URL \url{http://arxiv.org/abs/1904.10509}.

\bibitem[Dosovitskiy et~al.(2020)Dosovitskiy, Beyer, Kolesnikov, Weissenborn, Zhai, Unterthiner, Dehghani, Minderer, Heigold, Gelly, Uszkoreit, and Houlsby]{DBLP:journals/corr/abs-2010-11929}
Alexey Dosovitskiy, Lucas Beyer, Alexander Kolesnikov, Dirk Weissenborn, Xiaohua Zhai, Thomas Unterthiner, Mostafa Dehghani, Matthias Minderer, Georg Heigold, Sylvain Gelly, Jakob Uszkoreit, and Neil Houlsby.
\newblock An image is worth 16x16 words: Transformers for image recognition at scale.
\newblock \emph{CoRR}, abs/2010.11929, 2020.
\newblock URL \url{https://arxiv.org/abs/2010.11929}.

\bibitem[Vaswani et~al.(2017)Vaswani, Shazeer, Parmar, Uszkoreit, Jones, Gomez, Kaiser, and Polosukhin]{DBLP:journals/corr/VaswaniSPUJGKP17}
Ashish Vaswani, Noam Shazeer, Niki Parmar, Jakob Uszkoreit, Llion Jones, Aidan~N. Gomez, Lukasz Kaiser, and Illia Polosukhin.
\newblock Attention is all you need.
\newblock \emph{CoRR}, abs/1706.03762, 2017.

\bibitem[Carion et~al.(2020)Carion, Massa, Synnaeve, Usunier, Kirillov, and Zagoruyko]{DBLP:journals/corr/abs-2005-12872}
Nicolas Carion, Francisco Massa, Gabriel Synnaeve, Nicolas Usunier, Alexander Kirillov, and Sergey Zagoruyko.
\newblock End-to-end object detection with transformers.
\newblock \emph{CoRR}, abs/2005.12872, 2020.
\newblock URL \url{https://arxiv.org/abs/2005.12872}.

\bibitem[Devlin et~al.(2018)Devlin, Chang, Lee, and Toutanova]{DBLP:journals/corr/abs-1810-04805}
Jacob Devlin, Ming{-}Wei Chang, Kenton Lee, and Kristina Toutanova.
\newblock {BERT:} pre-training of deep bidirectional transformers for language understanding.
\newblock \emph{CoRR}, abs/1810.04805, 2018.
\newblock URL \url{http://arxiv.org/abs/1810.04805}.

\bibitem[Wang et~al.(2018)Wang, Girshick, Gupta, and He]{Wang_2018_CVPR}
Xiaolong Wang, Ross Girshick, Abhinav Gupta, and Kaiming He.
\newblock Non-local neural networks.
\newblock In \emph{Proceedings of the IEEE Conference on Computer Vision and Pattern Recognition (CVPR)}, June 2018.

\bibitem[Brown et~al.(2020)Brown, Mann, Ryder, Subbiah, Kaplan, Dhariwal, Neelakantan, Shyam, Sastry, Askell, Agarwal, Herbert{-}Voss, Krueger, Henighan, Child, Ramesh, Ziegler, Wu, Winter, Hesse, Chen, Sigler, Litwin, Gray, Chess, Clark, Berner, McCandlish, Radford, Sutskever, and Amodei]{DBLP:journals/corr/abs-2005-14165}
Tom~B. Brown, Benjamin Mann, Nick Ryder, Melanie Subbiah, Jared Kaplan, Prafulla Dhariwal, Arvind Neelakantan, Pranav Shyam, Girish Sastry, Amanda Askell, Sandhini Agarwal, Ariel Herbert{-}Voss, Gretchen Krueger, Tom Henighan, Rewon Child, Aditya Ramesh, Daniel~M. Ziegler, Jeffrey Wu, Clemens Winter, Christopher Hesse, Mark Chen, Eric Sigler, Mateusz Litwin, Scott Gray, Benjamin Chess, Jack Clark, Christopher Berner, Sam McCandlish, Alec Radford, Ilya Sutskever, and Dario Amodei.
\newblock Language models are few-shot learners.
\newblock \emph{CoRR}, abs/2005.14165, 2020.
\newblock URL \url{https://arxiv.org/abs/2005.14165}.

\bibitem[Britz et~al.(2017)Britz, Goldie, Luong, and Le]{britz-etal-2017-massive}
Denny Britz, Anna Goldie, Minh-Thang Luong, and Quoc Le.
\newblock Massive exploration of neural machine translation architectures.
\newblock In \emph{Proceedings of the 2017 Conference on Empirical Methods in Natural Language Processing}, pages 1442--1451, Copenhagen, Denmark, September 2017. Association for Computational Linguistics.
\newblock \doi{10.18653/v1/D17-1151}.
\newblock URL \url{https://aclanthology.org/D17-1151}.

\bibitem[Hu et~al.(2019)Hu, Zhang, Xie, and Lin]{DBLP:journals/corr/abs-1904-11491}
Han Hu, Zheng Zhang, Zhenda Xie, and Stephen Lin.
\newblock Local relation networks for image recognition.
\newblock \emph{CoRR}, abs/1904.11491, 2019.
\newblock URL \url{http://arxiv.org/abs/1904.11491}.

\bibitem[Ramachandran et~al.(2019)Ramachandran, Parmar, Vaswani, Bello, Levskaya, and Shlens]{DBLP:journals/corr/abs-1906-05909}
Prajit Ramachandran, Niki Parmar, Ashish Vaswani, Irwan Bello, Anselm Levskaya, and Jonathon Shlens.
\newblock Stand-alone self-attention in vision models.
\newblock \emph{CoRR}, abs/1906.05909, 2019.
\newblock URL \url{http://arxiv.org/abs/1906.05909}.

\bibitem[Zhao et~al.(2020)Zhao, Jia, and Koltun]{DBLP:journals/corr/abs-2004-13621}
Hengshuang Zhao, Jiaya Jia, and Vladlen Koltun.
\newblock Exploring self-attention for image recognition.
\newblock \emph{CoRR}, abs/2004.13621, 2020.
\newblock URL \url{https://arxiv.org/abs/2004.13621}.

\bibitem[Cordonnier et~al.(2019)Cordonnier, Loukas, and Jaggi]{DBLP:journals/corr/abs-1911-03584}
Jean{-}Baptiste Cordonnier, Andreas Loukas, and Martin Jaggi.
\newblock On the relationship between self-attention and convolutional layers.
\newblock \emph{CoRR}, abs/1911.03584, 2019.
\newblock URL \url{http://arxiv.org/abs/1911.03584}.

\bibitem[Ba~Alawi et~al.(2021)Ba~Alawi, Saeed, Almashhor, Al-Shathely, and Hassan]{9493430}
Abdulfattah~E. Ba~Alawi, Ahmed Y.~A. Saeed, Fatima Almashhor, Reem Al-Shathely, and Ahmed~N. Hassan.
\newblock Solid waste classification using deep learning techniques.
\newblock In \emph{2021 International Congress of Advanced Technology and Engineering (ICOTEN)}, pages 1--5, 2021.
\newblock \doi{10.1109/ICOTEN52080.2021.9493430}.

\bibitem[Wang et~al.(2020)Wang, Li, Dang, Ko, Han, and Moon]{article4}
Hanxiang Wang, Yanfen Li, L.~Minh Dang, Jaesung Ko, Dongil Han, and Hyeonjoon Moon.
\newblock Smartphone-based bulky waste classification using convolutional neural networks.
\newblock \emph{Multimedia Tools and Applications}, 79, 10 2020.
\newblock \doi{10.1007/s11042-020-09571-5}.

\bibitem[Dewulf(2017)]{phdthesis}
Victor Dewulf.
\newblock \emph{Application of machine learning to waste management: identification and classification of recyclables}.
\newblock PhD thesis, 10 2017.

\bibitem[Krizhevsky et~al.(2012)Krizhevsky, Sutskever, and Hinton]{article5}
Alex Krizhevsky, Ilya Sutskever, and Geoffrey Hinton.
\newblock Imagenet classification with deep convolutional neural networks.
\newblock \emph{Neural Information Processing Systems}, 25, 01 2012.
\newblock \doi{10.1145/3065386}.

\bibitem[Simonyan and Zisserman(2014)]{article6}
Karen Simonyan and Andrew Zisserman.
\newblock Very deep convolutional networks for large-scale image recognition.
\newblock \emph{arXiv 1409.1556}, 09 2014.

\bibitem[Szegedy et~al.(2014)Szegedy, Liu, Jia, Sermanet, Reed, Anguelov, Erhan, Vanhoucke, and Rabinovich]{DBLP:journals/corr/SzegedyLJSRAEVR14}
Christian Szegedy, Wei Liu, Yangqing Jia, Pierre Sermanet, Scott~E. Reed, Dragomir Anguelov, Dumitru Erhan, Vincent Vanhoucke, and Andrew Rabinovich.
\newblock Going deeper with convolutions.
\newblock \emph{CoRR}, abs/1409.4842, 2014.
\newblock URL \url{http://arxiv.org/abs/1409.4842}.

\bibitem[Szegedy et~al.(2015)Szegedy, Vanhoucke, Ioffe, Shlens, and Wojna]{DBLP:journals/corr/SzegedyVISW15}
Christian Szegedy, Vincent Vanhoucke, Sergey Ioffe, Jonathon Shlens, and Zbigniew Wojna.
\newblock Rethinking the inception architecture for computer vision.
\newblock \emph{CoRR}, abs/1512.00567, 2015.
\newblock URL \url{http://arxiv.org/abs/1512.00567}.

\bibitem[P et~al.(2021)P, Yadav, Shanmugam, V, and Suresh]{inproceedings}
Mallikarjuna P, Sachin Yadav, Aditi Shanmugam, Hima V, and Niharika Suresh.
\newblock Waste classification and segregation: Machine learning and iot approach.
\newblock pages 233--238, 04 2021.
\newblock \doi{10.1109/ICIEM51511.2021.9445289}.

\bibitem[Teh(2020)]{teh2020household}
Junjie Teh.
\newblock \emph{Household Waste Segregation Using Intelligent Vision System}.
\newblock PhD thesis, UTAR, 2020.

\bibitem[Xie et~al.(2016)Xie, Girshick, Doll{\'{a}}r, Tu, and He]{DBLP:journals/corr/XieGDTH16}
Saining Xie, Ross~B. Girshick, Piotr Doll{\'{a}}r, Zhuowen Tu, and Kaiming He.
\newblock Aggregated residual transformations for deep neural networks.
\newblock \emph{CoRR}, abs/1611.05431, 2016.
\newblock URL \url{http://arxiv.org/abs/1611.05431}.

\bibitem[Kumar et~al.(2021)Kumar, Buelaevanzalina, et~al.]{kumar2021efficient}
AP~Siva Kumar, K~Buelaevanzalina, et~al.
\newblock An efficient classification of kitchen waste using deep learning techniques.
\newblock \emph{Turkish Journal of Computer and Mathematics Education (TURCOMAT)}, 12\penalty0 (14):\penalty0 5751--5762, 2021.

\bibitem[Andhy Panca~Saputra(2021)]{andhy2021waste}
Kusrini Andhy Panca~Saputra.
\newblock Waste object detection and classification using deep learning algorithm: Yolov4 and yolov4-tiny.
\newblock \emph{Turkish Journal of Computer and Mathematics Education (TURCOMAT)}, 12\penalty0 (14):\penalty0 1666--1677, 2021.

\bibitem[Castellano et~al.(2019)Castellano, Carolis, Macchiarulo, and Rossano]{inproceedings2}
Giovanna Castellano, Berardina Carolis, Nicola Macchiarulo, and Veronica Rossano.
\newblock Learning waste recycling by playing with a social robot.
\newblock pages 3805--3810, 10 2019.
\newblock \doi{10.1109/SMC.2019.8914455}.

\bibitem[Náñez~Alonso et~al.(2021)Náñez~Alonso, Forradellas, Morell, and Jorge-Vázquez]{article7}
Sergio Náñez~Alonso, Ricardo Forradellas, Oriol Morell, and Javier Jorge-Vázquez.
\newblock Digitalization, circular economy and environmental sustainability: The application of artificial intelligence in the efficient self-management of waste.
\newblock \emph{Sustainability}, 13:\penalty0 2092, 02 2021.
\newblock \doi{10.3390/su13042092}.

\bibitem[Srinilta and Kanharattanachai(2019)]{Srinilta2019MunicipalSW}
Chutimet Srinilta and Sivakorn Kanharattanachai.
\newblock Municipal solid waste segregation with cnn.
\newblock \emph{2019 5th International Conference on Engineering, Applied Sciences and Technology (ICEAST)}, pages 1--4, 2019.
\newblock URL \url{https://api.semanticscholar.org/CorpusID:201066217}.

\bibitem[He et~al.(2015)He, Zhang, Ren, and Sun]{DBLP:journals/corr/HeZRS15}
Kaiming He, Xiangyu Zhang, Shaoqing Ren, and Jian Sun.
\newblock Deep residual learning for image recognition.
\newblock \emph{CoRR}, abs/1512.03385, 2015.
\newblock URL \url{http://arxiv.org/abs/1512.03385}.

\bibitem[Sandler et~al.(2018)Sandler, Howard, Zhu, Zhmoginov, and Chen]{8578572}
Mark Sandler, Andrew Howard, Menglong Zhu, Andrey Zhmoginov, and Liang-Chieh Chen.
\newblock Mobilenetv2: Inverted residuals and linear bottlenecks.
\newblock In \emph{2018 IEEE/CVF Conference on Computer Vision and Pattern Recognition}, pages 4510--4520, 2018.
\newblock \doi{10.1109/CVPR.2018.00474}.

\bibitem[Huang et~al.(2016)Huang, Liu, and Weinberger]{DBLP:journals/corr/HuangLW16a}
Gao Huang, Zhuang Liu, and Kilian~Q. Weinberger.
\newblock Densely connected convolutional networks.
\newblock \emph{CoRR}, abs/1608.06993, 2016.
\newblock URL \url{http://arxiv.org/abs/1608.06993}.

\bibitem[Sangi et~al.(2015)Sangi, Thompson, and Turuwhenua]{article8}
Mehrdad Sangi, Benjamin Thompson, and Jason Turuwhenua.
\newblock An optokinetic nystagmus detection method for use with young children.
\newblock \emph{IEEE Journal of Translational Engineering in Health and Medicine}, 3:\penalty0 1--10, 03 2015.
\newblock \doi{10.1109/JTEHM.2015.2410286}.

\bibitem[Thien(2012)]{Thien2012HorizontalGN}
Nicolas~Huynh Thien.
\newblock Horizontal gaze nystagmus detection in automotive vehicles.
\newblock 2012.
\newblock URL \url{https://api.semanticscholar.org/CorpusID:33586148}.

\bibitem[Li and Yang(2023)]{s23031592}
Haibo Li and Zhifan Yang.
\newblock Vertical nystagmus recognition based on deep learning.
\newblock \emph{Sensors}, 23\penalty0 (3), 2023.
\newblock ISSN 1424-8220.
\newblock \doi{10.3390/s23031592}.
\newblock URL \url{https://www.mdpi.com/1424-8220/23/3/1592}.

\bibitem[Pander et~al.(2012)Pander, Czabanski, Przybyla, Jezewski, Pojda-Wilczek, Wrobel, Horoba, and Bernys]{article9}
Tomasz Pander, Robert Czabanski, Tomasz Przybyla, Janusz Jezewski, Dorota Pojda-Wilczek, Janusz Wrobel, Krzysztof Horoba, and Marek Bernys.
\newblock A new method of saccadic eye movement detection for optokinetic nystagmus analysis.
\newblock \emph{Conference proceedings : ... Annual International Conference of the IEEE Engineering in Medicine and Biology Society. IEEE Engineering in Medicine and Biology Society. Conference}, 2012:\penalty0 3464--7, 08 2012.
\newblock \doi{10.1109/EMBC.2012.6346711}.

\bibitem[Turuwhenua et~al.(2014)Turuwhenua, Yu, Mazharullah, and Thompson]{article11}
Jason Turuwhenua, Tzu-Ying Yu, Zan Mazharullah, and Benjamin Thompson.
\newblock A method for detecting optokinetic nystagmus based on the optic flow of the limbus.
\newblock \emph{Vision Research}, 103, 10 2014.
\newblock \doi{10.1016/j.visres.2014.07.016}.

\bibitem[Haritosh et~al.(2019)Haritosh, Ralekar, Kaur, and Gandhi]{9029047}
Ankur Haritosh, Chetan Ralekar, Taranjit Kaur, and Tapan~Kumar Gandhi.
\newblock Human visual learning inspired effective training methods.
\newblock In \emph{2019 IEEE 16th India Council International Conference (INDICON)}, pages 1--4, 2019.
\newblock \doi{10.1109/INDICON47234.2019.9029047}.

\bibitem[Ba et~al.(2016)Ba, Kiros, and Hinton]{article13}
Jimmy Ba, Jamie Kiros, and Geoffrey Hinton.
\newblock Layer normalization.
\newblock 07 2016.

\end{thebibliography}






\end{document}